\documentclass[conference]{IEEEtran}
\IEEEoverridecommandlockouts
\usepackage{cite}
\usepackage{amsmath,amssymb,amsfonts}
\usepackage{listings}
\usepackage{algorithm}
\usepackage[noend]{algpseudocode}
\usepackage[english]{babel}
\usepackage[utf8]{inputenc}
\usepackage{graphicx}
\usepackage{textcomp}
\usepackage{xcolor}
\usepackage{url}
\usepackage{verbatim}
\usepackage{hyperref}
\usepackage{caption}
\usepackage{subcaption}

\def\BibTeX{{\rm B\kern-.05em{\sc i\kern-.025em b}\kern-.08em
    T\kern-.1667em\lower.7ex\hbox{E}\kern-.125emX}}
    
    \definecolor{lightGray}{gray}{0.9}
\begin{document}

\newcommand{\nb}[2]{
    \fbox{\bfseries\sffamily\scriptsize#1}
    {\sf\small\textcolor{red}{\textit{#2}}}
}
\newcommand\ag[1]{\nb{AG}{#1}}
\newcommand\tp[1]{\nb{TP}{#1}}
\newtheorem{puzzle}{Puzzle}

\title{Interleaving GANs with knowledge graphs to support design creativity for book covers}

\author{\IEEEauthorblockN{Alexandru Motogna and Adrian Groza}
\IEEEauthorblockA{\textit{Department of Computer Science} \\
\textit{Technical University of Cluj-Napoca}\\
Cluj-Napoca, Romania \\
\tt alexmotogna@gmail.com, adrian.groza@cs.utcluj.ro }


}

\maketitle

\begin{abstract}
An attractive book cover is important for the success of a book. 
In this paper, we apply Generative Adversarial Networks (GANs) to the book covers domain, using different methods for training in order to obtain better generated images. 
We interleave GANs with knowledge graphs to alter the input title to obtain multiple possible options for any given title, which are then used as an augmented input to the generator. 
Finally, we use the discriminator obtained during the training phase to select the best images generated with new titles. 
Our method performed better at generating book covers than previous attempts, and the knowledge graph gives better options to the book author or editor compared to using GANs alone.

\end{abstract}


\section{Introduction}
Cover of the book is the first impression on the reader. 
As such, it is important that it displays relevant depictions for the content of the book in an appealing manner in order to attract more readers. 
However, the process of designing and drawing the wanted book cover can be long and expensive. 
Our system aims to provide a much faster way to obtain the book cover and offers multiple versions, from which the author or editor may choose. 
Our system generates multiple book covers from a given title which can provide inspiration for the author or the artist to draw the final image. 
Our proposed method relies on interleaving Generative Adversarial Networks (GANs) with Knowledge Graphs (KGs). 

Aiming to include the human user in the loop (e.g., author, designer, editor), 
the developed system offers a choice of generated images, with the objective to generate more diverse images. 
For this we use a knowledge graphs to obtain similar and related words to the ones in the original title, to create new titles to be used as input to the generator aiming to create new and different options. Finally, to boost the quality of the images that are given to the human user, we use the discriminator trained with the GAN to determine the best looking images.

Our contributions are as follows:
First, we trained a conditional Generative Adversarial Network (i.e. the AttnGAN's architecture~\cite{attngan}) for generating book covers from an input title. 
Second, we improved the training by adding a set of technicalities: multi-GPU training, learning rate decay, discriminator training pause, Gaussian noise to discriminator's input and other tweaking of training parameters.
 Third, we used WordNet~\cite{wordnet} to obtain similar or relevant words to generate new titles that can then be used as new input titles for the generator, to get variety in the images returned by the system. 
 Fourth, we used the trained discriminator to select the best images out of those generated with the new titles to give the human agent a better selection.

\section{Related Work}

\subsection{Book Cover Generation}

Book cover synthesis from the summary has been done by Haque et al.~\cite{haque2022book}, which created a dataset of images with the summary of the book for training and used different generative models and evaluated the results. 
The text to image models used are AttnGAN \cite{attngan}, DF-GAN~\cite{DF-GAN}, both of which use a GAN architecture, and also DALL-E~\cite{ramesh2021dalle}, which uses a transformer-based approach. 
All three models generated covers do not have the same quality as that of an artist, as the art does not resemble real life object or scenery, but they do make a good impression. All models had a hard time reproducing text, which is almost always visible on book covers in the form of the title. 
To compare the models, two metrics for evaluating generative models have been used: Inception score~\cite{inceptionscore} and Frechet Inception Distance~\cite{frechetdistance}. 
DF-GAN scored the best for Inception Score and AttnGAN scored the best for Frechet Inception Distance. 
AttnGAN was able to mimic the structure of book covers the best as it always generated text for the title, although illegible. 
DF-GAN seemed to generate the best looking art, drawing recognizable human figures, and DALL-E generated random but very colorful images, but the quality of the art is subjective matter, so it is hard to choose the best out of the three.


Zhang et al.~\cite{zhang2021towards} have used an original approach to generate book covers with layout graphs. 
The layout graph is a structure for size, structure and positions of elements in the generated image. 
The proposed architecture is split into two modules: the layout generator and the title text generator. 
The layout generator uses Graph Convolution Networks to create the layout of the art and a GAN to generate the images. The title text generator adds the actual text of the input title above the image. 
This solution is very appropriate for generating book covers as it has a methodical approach. 
To our knowledge, no other methods for generating book covers have succeeded in having the title be readable on the final image. 
The generated images are very convincing book covers except for the fact that they are very blurry. 
Additionally, Zhang et al. have trained the generator for art on the COCO dataset~\cite{cocodataset}, a dataset of images of common objects, so the model is limited to only generating simple images. Together with the fact that it cannot learn any layouts from other book covers, the results are very plain, which is fitting for certain genres of books, such as cooking or science books, but less appropriate for genres such science fiction or mystery.

\subsection{Conditional Generative Adversarial Networks}

Generative Adversarial Networks, were first introduced by Goodfellow et al.~\cite{goodfellow2014generative} proposing the training of two models, a generator ($G$) and a discriminator ($D$). 
The generator learns the probability distribution over the data, while the discriminator learns the probability that a given image is from the original dataset. 
These two models are trained together and contribute to each other learning: images created by the generator are given to the discriminator to label and the quality of the generator is determined by how well the generated images have passed through the discriminator. 

The discriminator and the generator play a two-player Minmax game, where the discriminator tries to maximize $log(D(x))$ and the generator tries to minimize $log(1 - D(G(z)))$, where $D(x)$ is the discriminator's prediction that $x$ is from the original dataset and $G(z)$ is the data generated by the generator from a random noise $z$. The minmax game has a global optimum that the probability distribution of the dataset will be equal to the probability distribution learned by the generator and that the gradient descent training for GANs will converge to the global optimum.

Goodfellow has presented an unconditional GAN, where the generator only attempts at creating images that look like those from the dataset. 
Our task calls for a conditional GAN, where, besides trying to mimic the original images, the generator must follow a condition, in our case text, that changes how the image should look like. 
This can be achieved by adding the condition as an input to both the generator and the discriminator. 
Some conditional GAN architectures that performed well on different datasets and have been used in Haque et al.'s experiments to generate book covers~\cite{haque2022book}.

Tao et al. proposed a simpler approach to a GAN architecture with their Deep Fusion GAN~\cite{DF-GAN}. Tao et al argue that the strategy of stacking multiple generators, such as in the StackGAN~\cite{stackgan} or AttnGAN~\cite{attngan} architectures, has flaws, such as entanglement between generators, limited network supervision, and high computational cost. 
Since stacked architectures succeed in generating higher resolution images, the authors designed a Target-Aware Discriminator and a Deep text-image Fusion Block (DFBlock) to make up for the simpler network. The architecture also uses a One-Way Discriminator, which combines the conditional and unconditional loss into one value.
DF-GAN has very similar results on popular text-to-image datasets as other GAN architectures that are much more complex and in Haque et al.'s experiments it seems to generate the most recognizable objects in its art.


AttnGAN proposed by Xu et al.~\cite{attngan}, is a attention driven GAN architecture that is capable of generating images with finer detail by matching the areas of the image with the relevant words for it. As opposed to other GAN architecture they not only use features from sentences to condition the generation of the images, but also from words.

Similarly to StackGAN~\cite{stackgan}, which is the first attempt at stacking multiple generators, the AttnGAN network uses a similar stacking strategy that progressively increase the image size with each generator. The first generator is the most simple one, it mostly upsamples the sentence features and the noise. 
The next generator uses the output of the first one and and the output of an attentional model. 
This model computes how big of an impact words have on each subregion of the image. 
The following generators are identical, each using the output of the previous generator as input. 
In the depicted architecture only three generators are used because of GPU mempry constraints, but any number could be used. 
The final generator loss is computed by adding all the losses from all the generators, which increases when images are labeled as fake and when they do not match the sentence features, and the Deep Attentional Multimodal Similarity Model loss, a brand new addition to GAN architectures, which increases when images do not match the word features.

AttnGAN produces good results on popular text-to-image datasets, such as COCO~\cite{cocodataset}, having a much better inception score than previous GANs designed to that point. 
In Haque et al.'s experiments~\cite{haque2022book} AttnGAN followed the structure of a book cover the best out of all the architectures used, probably because the attentional models deal with subregions which promote the generation of a more structured image. Additionally the fact that word features are given a higher importance in the loss function is beneficial for book covers as it is to be expected that words in the title appear drawn on the cover.

\section{Training the GAN}

\subsection{Network architecture}

Of the many architectures used by Haque et al.~\cite{haque2022book} to generate book covers, the one used in our experiments is AttnGAN \cite{attngan} due to a few factors based on the results obtained by Haque et al.: scored the best for Frechet Inception Distance and a good result for Inception Score, the image quality was one of the best in our opinion and the network architecture seems to be the best for structured images due to its attention-based model~\cite{attngan}. 
DF-GAN's~\cite{DF-GAN} results seemed to have generated the better looking art but it did not always generate a spot for the title. 
DALL-E's~\cite{ramesh2021dalle} results looked the most confusing hardest to recognize. 
DALL-E does have the best results out of the three on a more generic dataset, such as Microsoft Common Objects in Context, but for book covers and a reduced training time and GPU count it struggles to produce the same quality of results.

The dataset on which the model has been trained on has been obtained by merging the dataset created and supplied by Haque et al.~\cite{haque2022book} and the dataset by~\cite{LukaKaggleDataset}, resulting in over 57.000 book covers labeled with their title.

\subsection{Improving training}
The first thing that was noticeable during training, especially with a large dataset, is that the training was very long, taking over two weeks. 
The original implementation of AttnGAN only supported running on one GPU which was very slow. 
By allowing training on multiple GPUs we can use bigger datasets, train for longer or do multiple training with different parameters in a shorter time. 
With this improvement, we were able to do the training in 4 days using 4 GPUs.

\begin{figure}
\centering
\includegraphics[scale=0.48]{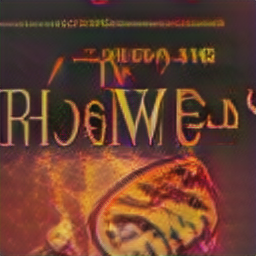}\hfill
\includegraphics[scale=0.48]{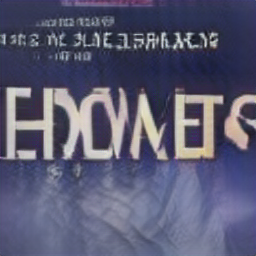}
\caption{Images generated from the title "Dragon Fire" (left) and "Dark Night" (right)}
\label{imgs2}
\end{figure}

\begin{figure}
\centering
\includegraphics[scale=0.5]{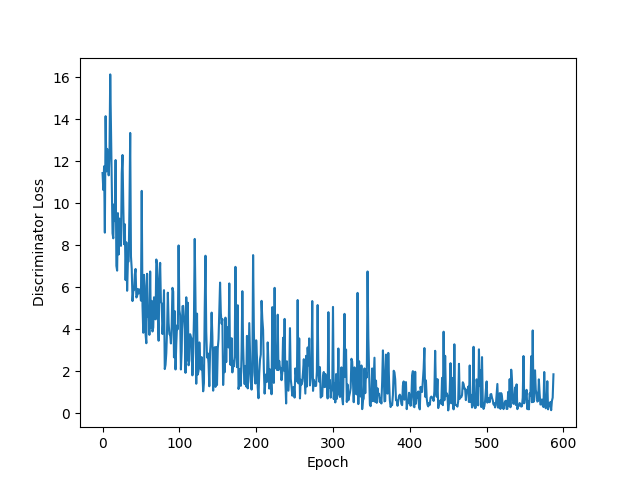}
\includegraphics[scale=0.5]{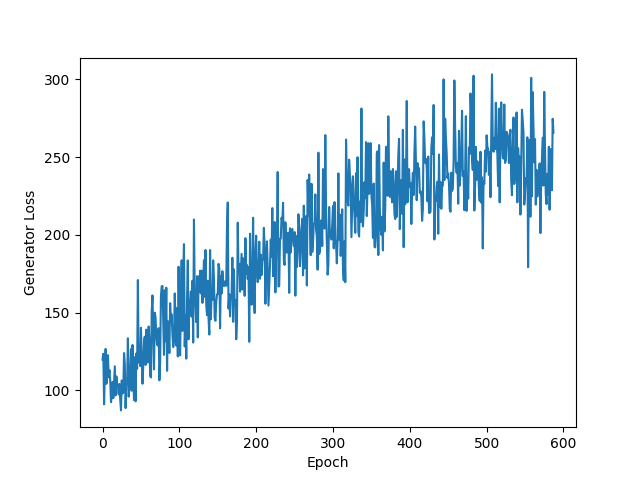}
\caption{Discriminator loss (on top) and Generator loss (on the bottom)}
\label{graphs2}
\end{figure}

Figure ~ref{imgs2} shows images that were obtained after training the GAN with the original AttnGAN architecture~\cite{attngan} with improvement of running on multiple GPUs. 
The image quality looks decent and some promising factors can be seen. First, it seems that the generated image always has an appropriate color scheme, and some relevant art is drawn. 
Second, the images follow the structure of the book cover well, showing text that is in good positions where the title and authors should be displayed. 
Unfortunately, that text is unrecognizable and unrepresentative of the title given as prompt, but a style of writing can be extracted as inspiration.

The more concerning problem of training is that looking at the loss graphs, shown in Figure~\ref{graphs2}, it can be seen that the loss value of the generator, which the model should be minimizing, is increasing at every epoch. 
Since the discriminator does seem to be converging, the problem is most likely the fact that the discriminator has an easier time learning, making it even harder for the generator to trick the discriminator into labeling the images wrongly and thus harder to improve.

In order to offset that, a few improvements have been developed. First, the learning rate of the generator has been decreased from an initial 0.002 to 0.0002. 
The learning rate is the training parameter that determines how much the model can change at each iteration. A high learning rate means that while changing it may overshoot the converging point, and a lower learning rate will help converge slowly but arrive closer to the wanted value. In order to customize the learning rate more, we have implemented learning rate decay. 
This way at any epoch intervals the learning rate will decrease a set amount. 
Hence, later into training when the generator is closer to the desired result, it will be able to easier to reach it, while at the start of training, changes can have a bigger impact since the model is far from its goal.

Aiming at giving the generator a needed advantage in training, we have also implemented an option to have the discriminator skip training a configurable amount of epochs. 
The goal of this change is to give the generator time to learn before the discriminator becomes too good at labeling fake images. 
We also used in our experiments the One-Way discriminator idea presented in DF-GAN~\cite{DF-GAN}, with the same goal of relaxing the discriminator a little.

\begin{figure}
\centering
\includegraphics[scale=0.6]{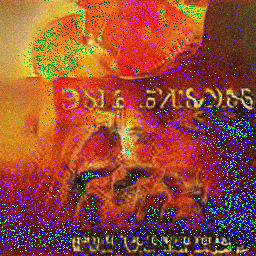}
\caption{Example of a generated image modified with Gaussian noise}
\label{noiseImage}
\end{figure}

Another approach to offset the distance between the generator and discriminator is to add Gaussian noise to every image that the discriminator is labeling during training. 
To compute the loss, the discriminator labels both real and fake images(fake images are images that were obtained from the generator). Both real and fake images get Gaussian noise in order to confuse the discriminator a little by making both types of images look a little worse and harder to distinguish. An example of such image can be seen in Figure~\ref{noiseImage}.

\section{Generating book covers with knowledge graphs}
For this part, we use the best generator and discriminator that we obtained through training. The goal is to offer multiple generated images to the  human agent, but not by using the same input multiple times, as that may lead to very similar results.

Our idea is to use a knowledge graph to get related or similar words, replace them in the original title, and obtain new titles that are then used as input to generate different images. 
This helps the generator explore other options with new words when generating. 
This can help the final result contain new depictions that the generator would not have yielded with the original title (e.g., on a book that has the wording "cooking" in its title, you might expect depictions of food on the book cover).

We use WordNet to find related words to create new titles. WordNet~\cite{wordnet} is a lexical database for English that groups words into synsets, which are groups of words that have the same meaning (synonyms). 
These words we look for are synonyms, hyponyms, hypernyms, and co-hyponyms of the synsets of the original word. 
Hyponyms are words with a more specific meaning(for example "herbivore" is a hyponym of "animal") and hypernyms are the opposite, a word with a broader meaning. Co-hyponyms are words that are hyponyms to the same synset(for example "dog" and "wolf" are co-hyponyms to the synset "canine"). To obtain a word's co-hyponyms we determine the hyponyms of all of it's hypernyms.

We get these related words for every word that is not a preposition or pronoun in the input title. As an additional improvement, we only use words that are in the vocabulary of our dataset, as we determined through testing that the generator performs very poorly on words it has never seen before. Then we perform random combinations by replacing the original words with the new words. For example, for the title "Adventure in a forest", we create titles such as "Chance in a wood" or "Hazard in a timber". Although these titles make no sense, their goal is to give the generator a more diverse input to obtain more diverse output images. 
The process of obtaining new titles is formalised in Algorithm~\ref{newWordsAlgo}.

\begin{algorithm}
\caption{Pseudo-code algorithm for obtain new titles}
\label{newWordsAlgo}
\begin{algorithmic}
    \Procedure{GetRelatedWords}{$word$}
        \State $newWords = []$
        \State $newWords.append(synonyms(word))$
        \For{each synsets in synsets(word)}
            \State $newWords.append(synset.hyponyms())$
            \State $newWords.append(synset.hypernyms())$
            \For{each hypernym in (synset.hypernyms()}
                \State $newWords.append(hypernym.hyponyms())$
            \EndFor
        \EndFor
        \State $removeDuplicates(newWords)$
        \State $removeWordsNotInDataset(newWords)$
        \State \textbf{return} newWords
    \EndProcedure

    \Procedure{GenerateNewTitles}{$title, number$}
        \State $newWordsMap = [][]$
        \For{each index, word in title}
            \If{word is not preposition or pronoun}
                \State $newWords = GetRelatedWords(word)$
                \State $newWordsMap[index].append(newWords)$
            \EndIf
        \EndFor
        \State $newTitles = []$
        \For{index in range(number)}
            \State $newTitle = ""$
            \For{words in newWordsMap}
                \If{$size(words[index]) \leq index$}
                    \State $newTitle.append(words[0])$
                \Else 
                    \State $newTitle.append(words[index])$
                \EndIf
            \EndFor
            \State $newTitles.append(newTitle)$
        \EndFor
        \State \textbf{return} newTitles
    \EndProcedure
\end{algorithmic}
\end{algorithm}

For the last part, we take the images generated with the new titles and pick only the ones that score the best for the unconditional score of the discriminator that was trained with the GAN. 
We use the unconditional one since it would not be fair to compare their conditional score for different input conditions. This step is important in order to avoid giving the author the generated images that have very poor quality, as our generator produces a significant amount of them, especially when give very unusual titles, such as the ones created with the related words. The image generated with the original title is always shown to the author, even if the quality is bad.

\begin{figure}
\centering
\includegraphics[scale=0.6]{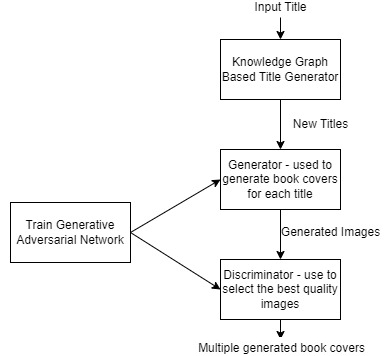}
\caption{The flow of the system. A GAN is trained to obtain a Generator and a Discriminator. A Knowledge Graph Based Title Generator creates new titles based on the input title. The titles are then given to the text-to-image generator to create book covers. The Discriminator selects the best images.}
\label{flowDiagram}
\end{figure}

Figure~\ref{flowDiagram} shows a flow diagram summarizing the process described in this paper.

\section{Experiments}
We trained the GAN in multiple ways aiming to successfully converging the model.
However, the loss shapes obtained in Figure~\ref{graphs2} indicate that the generator's loss is not decreasing.

Apart from evaluating the training of the GAN with the loss, we use Frechet Inception Distance (FID) \cite{frechetdistance} and Inception Score (IS) \cite{inceptionscore}. Both of these evaluation metrics are used to give an objective score to the quality of the generated images and were also used by Haque et al. \cite{haque2022book} to evaluate their experiments with generating book covers. 
We used the tool developed by Seitzer~\cite{fidcomputation} to compute the FID and~\cite{torchfidelity} to compute the IS. 
FID is a distance so a lower value is better, while for IS the bigger value is better.

Table~\ref{resultTable} lists the training experiments we have conducted with their respective parameters, results, IS and FID. 
Our models have obtained better for FID than in Haque et al.'s experiments on book covers~\cite{haque2022book} and better for IS when using the AttnGAN architecture~\cite{attngan}.

\begin{table*}
    \caption{Training versions of the GAN with their results}
    \label{resultTable}
    \centering
    \begin{tabular}{llllllll}
         No. & Generator LR & Generator LR Decay & Discriminator changes & Gaussian noise & IS & FID & Generated image example  \\
         \hline
         1 & 0.002 & None & Nothing & No & 4.09 & 31.36 & Figure \ref{imgs2} \\ 
         2 & 0.0002 & None & Nothing & No & 4.23 & 30.57 & Figure \ref{imgs3} \\
         3 & 0.0002 & Halves every 100 epochs & Skip training Every other epoch & No & 4.32 & 34.38 & Figure \ref{imgs4} \\
         4 & 0.0002 & Halves every 50 epochs & One-way Discriminator & No & 4.74 & 42.76 & Figure \ref{imgs5} \\
         5 & 0.0002 & None & Nothing & Yes & 4.66 & 42.26 & Figure \ref{imgs6}\\ 
         6 & 0.002 & None & Nothing & Yes & 4.29 & 42.43 & Figure \ref{imgs7} \\
    \end{tabular}
 \end{table*}

\begin{figure*}
\begin{subfigure}[b]{0.19\textwidth}
\centering
\includegraphics[scale=0.38]{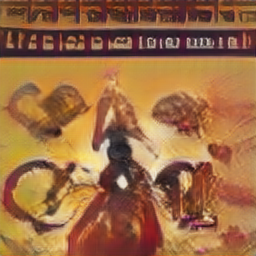}
\caption{"Dragon Fire"}
\label{imgs3}
\end{subfigure}\hfill
\begin{subfigure}[b]{0.19\textwidth}
\centering
\includegraphics[scale=0.38]{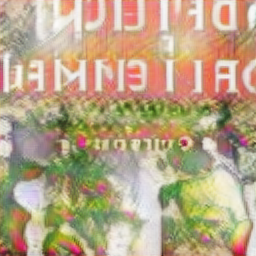}
\caption{"House in the garden"}
\label{imgs4}
\end{subfigure}\hfill
\begin{subfigure}[b]{0.19\textwidth}
\centering
\includegraphics[scale=0.38]{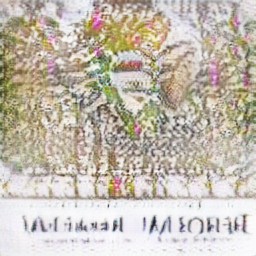}
\caption{"Adventure in a forest"}
\label{imgs5}
\end{subfigure}\hfill
\begin{subfigure}[b]{0.19\textwidth}
\centering
\includegraphics[scale=0.38]{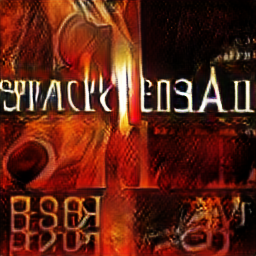}
\caption{"Dragon Fire"}
\label{imgs6}
\end{subfigure}\hfill
\begin{subfigure}[b]{0.19\textwidth}
\centering
\includegraphics[scale=0.38]{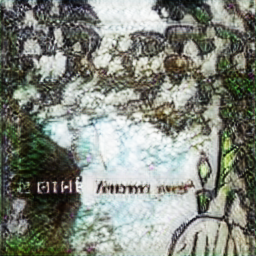}
\caption{"Adventure in a forest"}
\label{imgs7}
\end{subfigure}
\caption{Images generated using only GANs}
\end{figure*}

The quality of the pictures is not up to the standards of real book covers drawn by artists (Figure~\ref{fig:images}, but they still resemble the target decently.
We find that it consistently uses appropriate color schemes, create a unique structure, display text for the title in a fitting style, although illegible, and sometimes successfully generates recognizable objects that are relevant to the input title.

\begin{figure*}
\centering
\includegraphics[scale=0.35]{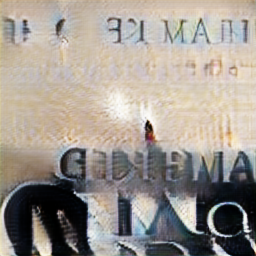}\hfill
\includegraphics[scale=0.35]{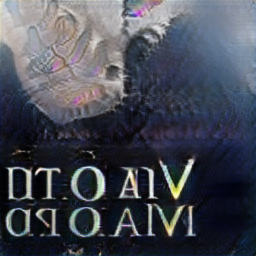}\hfill
\includegraphics[scale=0.35]{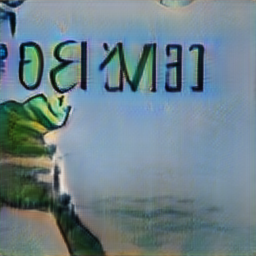}\hfill
\includegraphics[scale=0.35]{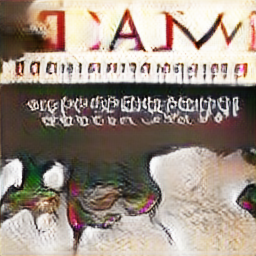}\hfill
\includegraphics[scale=0.35]{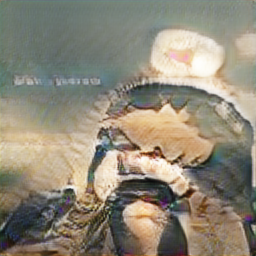}\hfill \vfill{0.5cm}

\includegraphics[scale=0.35]{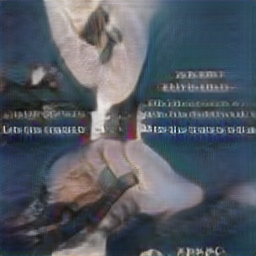}\hfill
\includegraphics[scale=0.35]{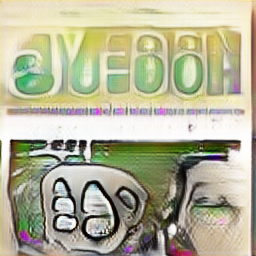}\hfill
\includegraphics[scale=0.35]{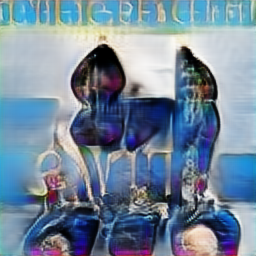}\hfill
\includegraphics[scale=0.35]{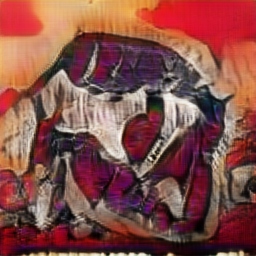}\hfill
\includegraphics[scale=0.35]{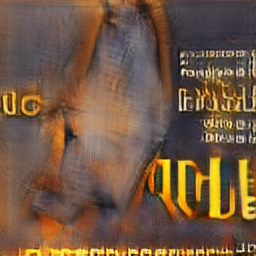}
\caption{Example of the generating pipeline with the input title "Lost at sea". 
The first image is the one generated with the original title. The next nine images (from left to right and then top to bottom) are the images generated with the title created with related words, ordered by unconditional score given by discriminator. The titles are in this order from left to right and then top to bottom: "Lost at Sea", "Bewildered at Sea", "Doomed at Ocean", "Missed at Tons", "Miss at Sea", "Baffled at Gulf", "Confused at Ford", "Lose at Bay", "Suffer at Stream", "Helpless at Ocean".}
\label{processExample}
\end{figure*}

Apart from the using the generator directly, we experimented using the entire pipeline for generating multiple book covers presented in this paper, to see if the process does deliver at least one image that is better than the one obtained directly from the generator with the title. An example of the results with the input title "Lost at Sea" is show in Figure~\ref{processExample}.

We argue that this process does in fact provide a better selection and even better choice than the original, as our favorite is the third picture generated with the title "Doomed at Ocean". 
It can also be seen that the quality does decrease with every image and the last 4 images would not be shown to the author, as they are the worse of the bunch.


\section*{Code availability}
The code for training is available at \url{https://github.com/AlexMotogna/AttnGAN}, a fork of the original AttnGAN repository \url{https://github.com/taoxugit/AttnGAN}.
The code for generating book covers with a pre-trained model for book covers is available at \url{https://github.com/AlexMotogna/GeneratorAPI}.

\section{Conclusion}

We used here various training techniques for Generative Adversarial Networks, such as learning rate decay and Gaussian noise, to train GANs on book cover by their title, and use a knowledge graph of semantics to create new titles which boost the diversity of the generated art. Finally, we use the discriminator to find the best looking images to give to the author.
The training techniques explored proved to be insufficient for the convergence of the GAN but did slightly increase the quality of the generated images. 
The pipeline for giving the human agent more options to choose from proves to be successful in giving a wider variety and delivering at least one better option than the original.

An improvement would be  to handle illegible text. This would ideally be done by training on a dataset that is exclusively book cover art, (i.e. without the text of the title or others) so that the generator can learn to create appropriate art for book covers while ignoring the text. 
In line with Zhang et al.~\cite{zhang2021towards},  the text could later be added. 

\bibliographystyle{IEEEtran}
\bibliography{bib} 

\end{document}